# Backtrack Tie-Breaking for Decision Trees:
## A Note on Deodata Predictors

**Cristian Alb**

CA.PUBLICUS@GMAIL.COM

### Abstract

A tie-breaking method is proposed for choosing the predicted class, or outcome, in a decision tree. The method is an adaptation of a similar technique used for deodata predictors.

## 1    Introduction

Decision trees make predictions about the outcome, or class label, of a query defined by a set of attribute values [1]. The choice of the predicted outcome is based on the count of the training outcomes associated to the corresponding node in the tree. The training outcome with the highest count is chosen as the predicted outcome. However, it can happen that more than one outcome matches the highest count. This is referred to as a tie situation. For such situations, the default action is to randomly pick one of the outcomes with the highest count [2].

A better method to break ties is suggested herein. It is an adaptation of the tie-breaking technique described in [3].

## 2    Backtrack Tie-Breaking (BTTB)

Tie-breaking mechanisms improve the accuracy of classifiers. This is an intuitive expectation that finds support in practical experiments. For instance, in article [3] there is a section on experiments that show a statistically significant improvement of accuracy when a tie-breaking mechanism is present. Two algorithms that differ only by the presence of such a tie-breaking mechanism are tested. The algorithms are referred to by the identifiers "deodata_delanga" and "deodata_tbreak_delanga". The corresponding accuracies are estimated at 0.601807831 and 0.615366088, respectively. The estimates are based on a number of 28,569,600 tests [3, p.13].

The tie-breaking mechanism of the deodata algorithm is based on a backtracking technique. The same technique can be adapted for decision tree classifiers. When building the decision tree, the outcomes associated to the evaluated nodes should be checked for ties. If more than one outcome shares the maximum number of occurrences, the tie breaking technique should be applied. The technique consists in going up to the parent node and examining the list of outcomes associated to it. The counts of the outcomes that were previously tied are accumulated. If, as a result of the addition of new counts, a winner emerges, that outcome is selected as the prediction. If a tie still persists, the algorithm should move up again to the parent of the parent node. The procedure is repeated while ties persist. If at the root of the decision tree a tie still persists, the predicted outcome is picked at random.

In Table 1 a trivial example of a training data set is shown. A decision tree implementing a classifier for the training data is shown in Fig. 1.

| Index | Outcome | Attr A | Attr B |
|-------|---------|--------|--------|
| 0 | t3 | a1 | b0 |
| 1 | t0 | a1 | b0 |
| 2 | t0 | a0 | b1 |
| 3 | t1 | a0 | b1 |
| 4 | t2 | a1 | b1 |
| 5 | t2 | a1 | b1 |
| 6 | t2 | a0 | b0 |
| 7 | t1 | a0 | b0 |

Table 1: Classifier training data set.

The non-leaf nodes are represented by circles. Inside the circle there is an indication of the attribute used to expand the branches. The trapezoid shape represents leaf nodes and contains the associated outcome values.

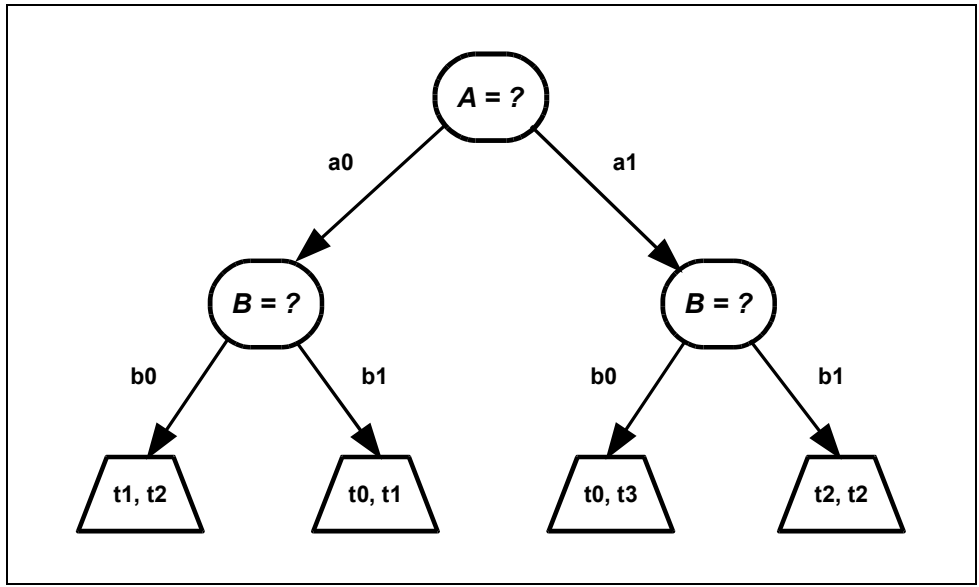

Figure 1: Classifier decision tree.

The procedure for the backtrack tie-breaking is illustrated in Fig. 2. The nodes of the decision tree are marked with labels from N1 to N7. Next to each node there is a table representing the outcomes associated to it, ordered by number of counts. The tables also contain, for each outcome value, a list of indexes corresponding to the entries in Table 1.

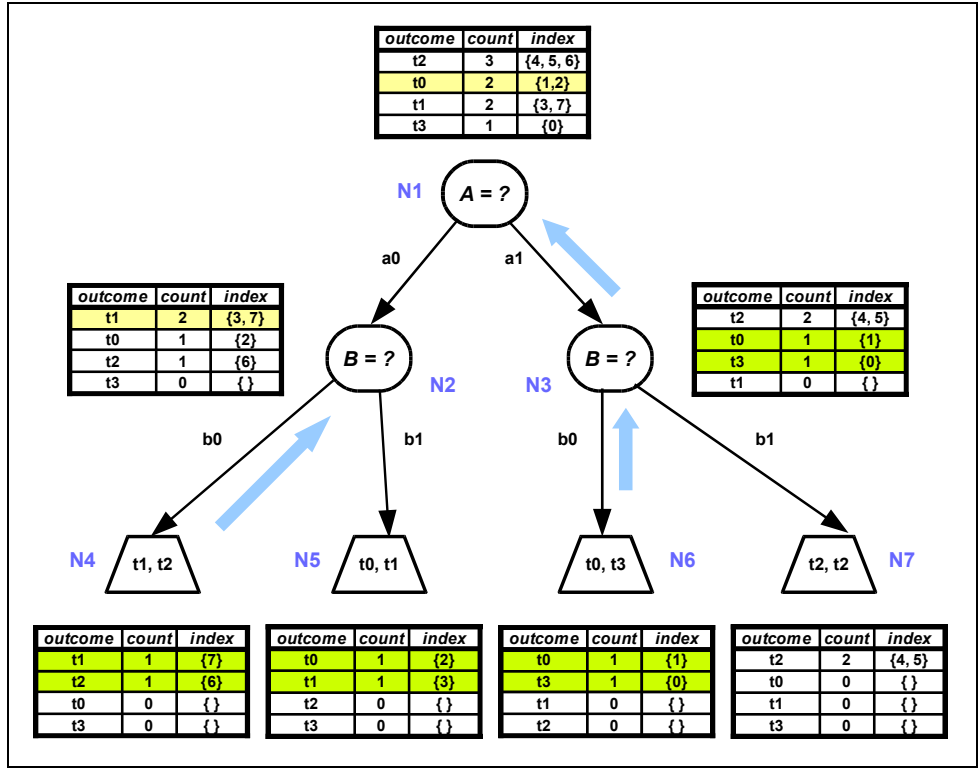

Figure 2: Tie-breaking procedure.

As an example, if considering the classification corresponding to the leaf node N4, the outcome values associated to it are 't1' and 't2'. The counts for the outcomes are tied at one each. Instead of picking at random, the focus should move to the parent node, N2. The counts associated to N2 for the tied outcomes are: two for 't1' and one for 't2'. Therefore, the tie is broken and 't1' is selected as the outcome/class label associated to node N4.

If considering the classification that corresponds to node N6, the outcome values associated to it are 't0' and 't3' with a count of one each. Focus is moved to to the parent node, N3. The counts associated to N3 for the tied outcomes are one each. The tie still persists. Once again, the focus is moved up to N1, the parent node of N3. The counts corresponding to the tied outcomes associated to N1 (the root of the tree) are: two for 't0' and one for 't3'. The tie is broken and 't0' is selected as the outcome/class label associated to node N6.